# Cooperative Causal GraphSAGE

Zaifa Xue[a,b], Tao Zhang[a,b], Tuo Xu[a,b], Huaixin Liang[a,b], Le Gao[a,b]

a School of Information Science and Engineering, Yanshan University, Qinhuangdao, China

b Hebei Key Laboratory of information transmission and signal processing, Qinhuangdao, China

*Abstract*—**GraphSAGE is a widely used graph neural network. The introduction of causal inference has improved its robust performance and named as Causal GraphSAGE. However, Causal GraphSAGE focuses on measuring causal weighting among individual nodes, but neglecting the cooperative relationships among sampling nodes as a whole. To address this issue, this paper proposes Cooperative Causal GraphSAGE (CoCa-GraphSAGE), which combines cooperative game theory with Causal GraphSAGE. Initially, a cooperative causal structure model is constructed in the case of cooperation based on the graph structure. Subsequently, Cooperative Causal sampling (CoCa-sampling) algorithm is proposed, employing the Shapley values to calculate the cooperative contribution based on causal weights of the nodes sets. CoCa-sampling guides the selection of nodes with significant cooperative causal effects during the neighborhood sampling process, thus integrating the selected neighborhood features under cooperative relationships, which takes the sampled nodes as a whole and generates more stable target node embeddings. Experiments on publicly available datasets show that the proposed method has comparable classification performance to the compared methods and outperforms under perturbations, demonstrating the robustness improvement by CoCa-sampling.**

## I. INTRODUCTION

Graph neural networks (GNNs) and graph representation learning have become effective tools for analyzing graph structure data in recent years, which have demonstrated excellent performance in the fields of citation network prediction [1], social network analysis [2], recommendation system [3], and molecular information discovery [4]. Early graph neural models focused on how to efficiently transfer and aggregate information over graph structures to learn the representation of nodes. Hamilton et al. designed GraphSAGE, an inductive graph neural network learning framework for large-scale graph data, to conduct a fixed number of random samples of neighborhood nodes to reduce the computational complexity, and to learn an aggregate function to generate an embedded representation for unfamiliar nodes, enhancing the generalization ability of the model [5]. In order to improve the learning efficiency of graph convolution on large-scale graphs, Chen et al. introduced an importance-based hierarchical sampling strategy to reduce the number of nodes involved in calculation and reduce the computational complexity of the model [6]. Veličković et al. introduced an attention mechanism to weight and aggregate the feature information of neighborhood nodes, so that the model can focus on important nodes adaptively and improve the flexibility of the model [7]. In subsequent studies, Oh et al. utilized reinforcement learning to infer the importance of neighborhood nodes, and sampled neighborhood nodes according to this standard, reducing the variance of GraphSAGE [8]. Zhao et al. learned the sampling method by combining the sampling process with forward propagation, which improved the accuracy while speeding up the convergence of the loss function [9]. Li et al. proposed a graph neural network framework based on course learning, which improved the performance in the case of imbalance in the node category labels [10]. Zhang et al. proposed an improved graph based on the introduction of external node convolutional neural network model to improve the accuracy of small sample node classification [11]. Dornaika et al. integrated graph learning and graph convolution into a unified architecture that adaptively weights nodes based on the distance of labeled nodes to category boundaries to improve semi-supervised node classification performance [12].

However, with the increasing complexity of application scenarios, graph neural networks have put forward higher requirements for the classification ability and robustness of models, and causal inference has been introduced into graph neural networks by some researchers, and has shown effective performance improvement. Zhang et al. proposed a causal sampling algorithm, which utilized the causal relationship between nodes and labels to calculate the causal effect of nodes, and used it as the sampling weight of neighborhood nodes to solve the problem of causal confusion of GraphSAGE and improve the robustness of the model [13]. Lee et al. also designed a causal graph for graph data, used the causal mechanism of graph convolutional networks to eliminate the influence of bias, and further divided the attention mechanism into node attention and neighborhood attention, which also improved the robustness of the model [14]. Liu et al. generated a causal graph between blood pressure changes and wearable features from the perspective of causal inference, and then utilized the graph neural network to learn from the causal graph to improve the accuracy of blood pressure measurement [15]. Zhai et al. developed a click-through rate prediction model under the graph neural network framework based on the causal relationship between the features in the graph, which improved the performance of the click-through rate prediction task [16]. Tan et al. introduced counterfactual and factual reasoning method for solving learning and evaluation problems in interpretable graph neural networks [17]. Although existing work has yielded some results, the cooperative relationship of neighboring nodes in aggregation process has received little attention. This may affect the ability of the model to capture local structural information, leading to suboptimal performance [18]. The complexity of graph data and the interrelationships between nodes require more perspectives to understand and optimize graph neural network models [19].

To address these issues, this paper proposes Cooperative Causal GraphSAGE, termed CoCa-GraphSAGE, a novel model that combines causal inference and cooperative relationships. In this model, cooperative causal sampling (CoCa-sampling) is adopted in the process of neighborhood node sampling. Specifically, the corresponding causal effects of nodes are calculated based on causal graphs constructed by considering node coalitions under different neighborhoods. For all eligible coalitions, cooperative causal weights can be obtained to guide the node sampling process. The target node embeddings are then generated by aggregating the features of the sampled nodes. To verify the performance of CoCa-GraphSAGE, the proposed model is compared with seven comparison models on five base node classification datasets. In addition, perturbation experiments are performed to verify the robustness of CoCa-GraphSAGE. The experimental results verify the effectiveness of CoCa-GraphSAGE. The main contributions of this paper are summarized as follows:

1) This paper proposes a graph neural network model integrating Causal GraphSAGE and cooperative game relationship, which makes full use of the cooperative relationships between neighboring nodes and causal information between nodes and labels, adjusts the sampling strategy of the GraphSAGE model, and thus improves the classification performance of the model.

2) In this paper, a cooperative causal sampling algorithm is designed. First, a cooperative causal structural model is constructed in the cooperative situation, and then Shapley values and causal effect method are linked to calculate the cooperative causal weights of nodes, so as to solve the weight bias caused by the neglected node in the causal sampling.

The remainder of this paper is organized as follows. In Section II, this paper reviews the researches on Causal GraphSAGE and cooperative game theory on graph networks, and discusses how the previous researches inspired the work in this paper. In Section III, the unsolved problems are analyzed and the proposed CoCa-GraphSAGE model is described in detail. The experimental results and analysis are presented in Section IV. Section V summarizes the content of this paper and draws a conclusion.

## II. RELATED WORKS

In this section, we briefly review the researches on Causal GraphSAGE and cooperative game theory on graph networks, and explain the connection between these researches and the method presented in this paper.

### A. Causal GraphSAGE

In order to solve the problem of causal confounding on GraphSAGE caused by perturbations, Zhang et al. introduced causal inference into GraphSAGE and proposed Causal GraphSAGE [13]. Firstly, Causal GraphSAGE(C-GraphSAGE) constructs a causal structure model based on the potential causal relationship. Fig. 1 shows the abstract causal model of C-GraphSAGE.

Secondly, according to the established causal graph, C-GraphSAGE uses causal sampling instead of conventional random sampling in GraphSAGE and calculates the causal weights of nodes according to the intervention formula. In the causal diagram of Fig.1, the relationship $v_i \leftarrow U \rightarrow y_i$ satisfies the backdoor criterion. When $v_n = v_i$ and $y_n = y_i$, the causal weight can be obtained according to the causal sampling principle, as shown in (1).

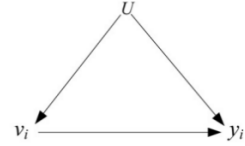

**Fig. 1.** The abstract causal graph underlying C-GraphSAGE.

$$p\left(y_i \mid do(v_i)\right) = \frac{1}{\hat{N}\hat{p}(v_i \mid U)} \tag{1}$$

Where $\hat{p}$ represents the estimate of $p$. In the causal sampling phase, for all candidate neighbors, their causal weights are calculated according to the above formula. Finally, nodes with stronger causal effects will be sampled.

However, this causal sampling strategy does not take into account the potential relationship between the neighborhood nodes, but simply abstracts the neighbor nodes into a whole. The research in this paper pays more attention to the cooperative relationship between different neighborhood nodes.

### B. Cooperative Game Theory

Cooperative game theory mainly studies the distribution of payoffs generated when multiple participants cooperate, with the purpose of finding a cooperation scheme that enables participants to obtain fair payoffs in cooperation [20]. Shapley values, derived from game theory, incorporate the potential for each collaborator to interact with other collaborators, thereby constructing a fair and efficient allocation scheme [21]. Specifically, in the neighborhood aggregation stage of node $v_i$, each node in the neighborhood node set is regarded as a player, and the Shapley value of a node can be calculated by (2), as follows:

$$\phi_j = \sum_{\mathcal{S} \subseteq \mathcal{N}(v_i) \backslash v_j} (v(\mathcal{S} \cup v_j) - v(\mathcal{S})) \frac{|\mathcal{S}|!(n-|\mathcal{S}|-1)!}{n!} \tag{2}$$

Where $v(\bullet)$ is the contribution function, used to measure the contribution degree of members to the goal; $n$ represents the total number of participants. By synthesizing the contributions of $v_j$ in various cooperation cases, the average value of the contributions of $v_j$ in the sampling and aggregation process can be obtained to reflect the average influence of $v_j$ on the prediction center node label.

Alexandre et al. proposed a method utilizing Shapley values to explain cooperation strategies in reinforcement learning, which can well evaluate the contribution of each member in the cooperation [22]. Sun et al. proposed the use of Shapley values to consider the interactions between features in the process of constructing a random forest for assessing the importance of each candidate feature [23]. Li et al. then pooled Shapley values with deep neural networks for analyzing the interactions between factors in a cooperative game model in order to improve the interpretability and prediction accuracy of the model [24]. All these studies have utilized the important role of Shapley values in the analysis of cooperative game relationships. In combination with graph neural networks, Narayanam et al. proposed an influence node algorithm based on Shapley values, which is used to find key nodes with the smallest size in social networks and improve the computational efficiency of information diffusion degree in social networks [25]. Li et al. utilized Shapley values to analyze and rank the importance of topological nodes in software-defined networks, providing an effective method for understanding and explaining the process of learning network topology and forecasting by graph neural network models [26]. Mastropietro et al. proposed an edge Shapley interpreter to explain the importance of edges in graph data to the prediction of graph neural network models, and applied it to the activity prediction task of compound molecules [27]. The above research results indicate that Shapley value is beginning to emerge in graph neural networks, but there is no research to combine it with causal graph neural networks.

## III. METHODOLOGY

In this section, firstly, the problems in causal sampling method are analyzed. Secondly, CoCa-GraphSAGE model is proposed and described in detail, mainly including cooperative causal model and CoCa-sampling. Fig. 2 shows the process structure diagram of CoCa-GraphSAGE obtaining an embedded representation of a node with a first-order neighborhood as an example. CoCa-GraphSAGE can be extended to multi-order neighbor nodes. An undirected graph is denoted by $G = (V, E)$, where $V = \{v_1, v_2, \cdots, v_N\}$ denotes the set of nodes. And the edge set is denoted by $E \subseteq V \times V$, where $(v_i, v_j) \in E$ denotes that there is an edge between $v_i$ and $v_j$, which means that $v_i$ and $v_j$ are neighboring vertices to each other. $Y = [y_1, y_2, \cdots, y_N]$ is the label vector of all nodes, where $y_i \in C$ denotes the label of $v_i$, with $C$ being the label set. For $\forall v_r \in V$, the first-order neighborhood is expressed as $N(v_r) = \{v_k \mid v_k \in V \wedge (v_k, v_r) \in E, v_k \neq v_r\}$. Specifically, we firstly determine all potential combinations of cooperative alliances among the neighboring nodes. These alliances are then utilized to calculate the cooperative causal weights corresponding to each of the neighboring nodes. For all candidate neighbors, the weights from their own features to their respective labels are computed. Subsequently, the neighbor nodes with relatively higher weights are selected for sampling. The information of the sampled nodes is then aggregated to update their own representations, facilitating accurate label prediction for node classification.

### A. Problem Statement

Before conducting causal sampling, it is necessary to build a causal structure model in the graph by utilizing causal relationship. In this process, all neighboring nodes are considered as confounding factors that have an impact on causality, except for the nodes whose weights are to be calculated. Based on this setting, the causal weights of nodes for sampling are calculated. However, once the sampling process is completed, the unsampled neighborhood node will no longer participate in the subsequent information aggregation of the central node, resulting in the change of the information source of the central node and the change of the causal relationship between the neighbors. The change of causality leads to the change of confounding factors in the causality structure graph, which makes the original causal weight calculated based on the old causal structure no longer applicable to the updated causal structure, leading to the weight bias. This misjudgment not only reduces the ability of sampling to identify and capture the causal relationship in the graph data, but also affects the learning ability of model and robustness to the graph data.

### B. Cooperative Causal Model

To further improve the robustness of GraphSAGE, this paper utilizes CoCa-sampling as the sampling strategy. First, the cooperative causal structure graph is established, as shown in Fig. 3. This section describes in detail the establishment process of the model.

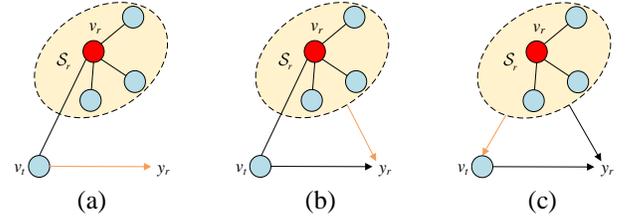

**Fig. 3.** The cooperative causal structure graph of CoCa-GraphSAGE. (a)The causality between neighborhood nodes and labels. (b)The causality between sampling sets and labels. (c)The causality between nodes.

Before sampling, all nodes in the neighborhood node set are alternative nodes, which act as components of players to form a gaming coalition to participate in the cooperative gaming process. For the node $v_r$ in the graph and its neighboring node set $N(v_r)$, the cooperative causal analysis method focuses on studying the causal relationship between a node $v_i$ in $N(v_r)$ and the label $y_r$ of $v_r$ in the process of feature aggregation generation. Under the action of the aggregation function, the characteristics of the node $v_i$ will be aggregated into the node $v_r$ to generate its representation, and used to predict the label $y_r$, that is, $v_i$ is one of the reasons for the label $y_r$. Therefore, there is a causal path $v_i \rightarrow y_r$ in the causal model, which is used to reflect the causal effect of the neighborhood node on the central node label, as shown by the orange line at the bottom of Fig. 3(a).

Secondly, in order to get close to the real sampling environment, a neighborhood node set $\mathcal{S} = \mathcal{N}(v_r) \backslash v_r$, which cooperates with node $v_r$, is introduced, so $v_r$ and $\mathcal{S}$ together constitute the sampling node set. The feature information of any node $v_j$ in $\mathcal{S}$ will be used to generate the embedded representation of the node $v_r$, which also includes the feature information of the node $v_r$ itself. Therefore, both $\mathcal{S}$ and $v_r$ are the reason of label $y_r$. And if the two are represented as $\mathcal{S}_r = \mathcal{S} \bigcup v_r$, there is still a causal path $\mathcal{S}_r \to y_r$ in the causal model, as shown by the orange line on the right in Fig. 3(b).

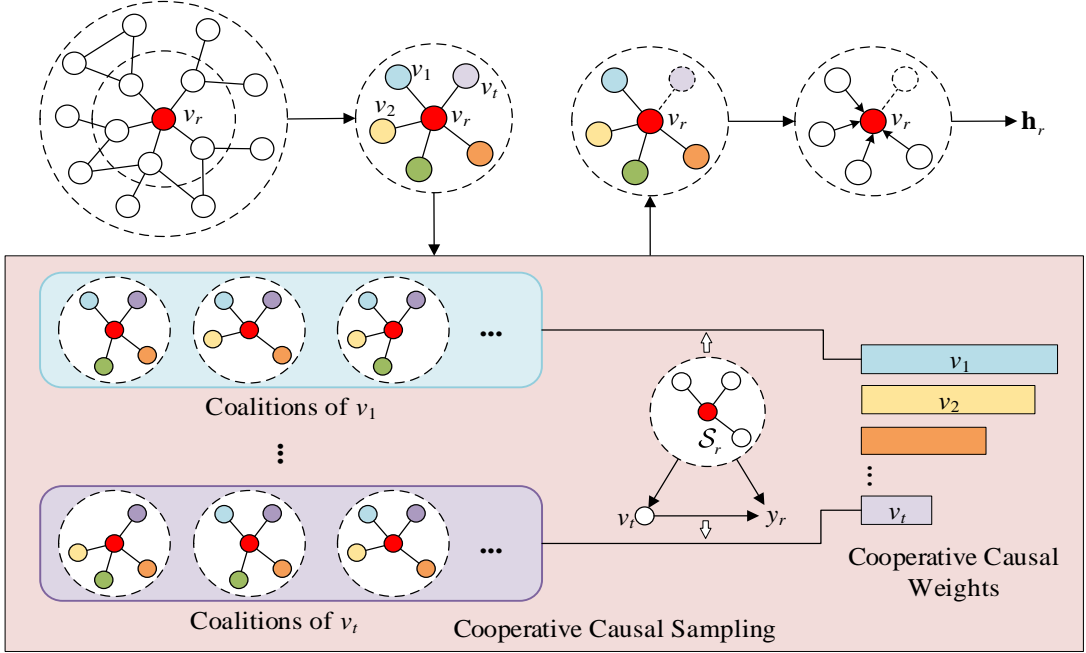

**Fig. 2.** The framework of CoCa-GraphSAGE on first-order neighbor nodes. The neighborhood node weights in CoCa-sampling are obtained as shown on the top. Different neighborhood nodes are distinguished by different colors, and different rectangle lengths represent the sampling weights of neighborhood nodes in CoCa-sampling.

At the same time, from the perspective of aggregated information, for any node $v_j \in \mathcal{S}$, there is always an edge $< v_t, v_r, v_j >$ in the graph. In the process of information propagation of feature aggregation in the graph, the information of the node $v_r$ and the node set $\mathcal{S}$ will be transmitted to the node $v_t$ through this side path, that is to say, the information in $\mathcal{S}_r$ will affect the node $v_t$. Therefore, there is a causal path $\mathcal{S}_r \to v_t$ in the process of aggregation, as shown by the orange line on the left in Fig. 3(c). Based on the above analysis, a cooperative causal structure model containing causal information and cooperative relationships between neighboring nodes is constructed on the graph under certain sampling conditions.

However, considering the data perturbation existing in the real environment, so it is also necessary to construct the cooperative causal model in the case of perturbations, as shown in Fig. 4.

Firstly, the perturbations will directly or indirectly affect the node information, structure information and label information in the graph, so there is a causal relationship between the perturbation and $v_t$, $\mathcal{S}_r(v_t)$ and $y_r$. *dis* represents external perturbation, that is, there are three causal paths: $dis \to y_r$, $dis \to v_t$, $dis \to \mathcal{S}_r$, as shown in Fig. 4(a).

In order to highlight and analyze the causal relationship between $v_t$ and $y_r$, the cooperative causal model under the influence of perturbation is simplified by using $D_{\mathcal{S}}$ to denote the perturbation and the cooperative neighborhood nodes affected by the perturbation, as shown in Fig. 4(b).

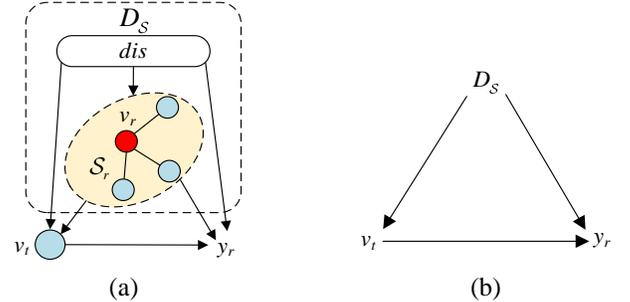

**Fig. 4.** The cooperative causal model considering the influence of perturbations. (a) Cooperative causal model considering perturbations. (b) Simplified cooperative causal model.

### C. Cooperative Causal Sampling Algorithm

Here, the case without perturbation is taken as an example to calculate cooperative causal weight. In Fig. 3, the existence of $\mathcal{S}_r$ leads to the existence of the causal path $v_t \leftarrow \mathcal{S}_r \to y_t$ between $v_t$ and $y_r$, and this path meets the backdoor criterion. According to the description of C-GraphSAGE, when considering that the node $v_t$ have been selected to participate in the label prediction, the causal weight of $v_t$ on the label $y_r$ can be expressed as (1). This causal weight can be thought of as the payoff of the node. Considering that causal sampling only considers the weights of individuals for node sampling, this paper establishes the cooperative causal structure model in the case of cooperation, so the payoff of the coalition set composed of different nodes should also be considered when calculating causal weights. For a definite coalition of nodes $S_r$, the causal weight is expressed as follows:

$$p\left(y_r \mid do(v_t)\right)\Big|_{S_r} = \frac{1}{N\hat{p}\left(v_t \mid S_r\right)} \qquad (3)$$

Where, $\hat{p}$ denotes the kernel density estimate of $p$, and the above equation reflects the causal weight of node $v_t$ on the predicted label $y_r$ when the subset of neighborhood nodes is the subset $S_r$ of cooperative game, which is the marginal causal weight. As shown in Fig. 2, the set of possible node coalitions for a certain node $v_t$ is not unique. The sets of all node coalitions are defined as $S_1, S_2, \cdots, S_r, \cdots, S_j \subseteq N(v_r) \setminus v_t$, so the corresponding causal weight is $p(y_r \mid do(v_t))\big|_{S_j}$. In order to comprehensively consider the causal weight of $v_t$ in the cooperative process, we define cooperation expectation to combine all causal weights. By referring to the Shapely values formula, this paper combines the causal weight with the marginal contribution, and the marginal contribution $\Delta(S_r)$ of the set of a certain coalition $S_r$ at a certain node $v_t$ can be obtained as follows:

$$\Delta(S_r) = p\left(y_r \mid do(v_t)\right)\Big|_{S_r \cup v_r} - p\left(y_r \mid do(v_t)\right)\Big|_{S_r} \qquad (4)$$

Secondly, for node sampling, let sampling number to be $M$ and then there is $\left|S_j \cup \{v_t\}\right| = M$. In this case, the probability of all node coalitions can be calculated to expand the cooperative causal weight, namely:

$$\mathbb{E}(v_r) = \sum_{S_j} \frac{(M-1)!(T-M)!}{T!} \Delta(S_j) \qquad (5)$$
$$s.t. S_j \subseteq N(v_r) \setminus v_t, |S_j| = M-1$$

Substituting the marginal contribution formula into (5), the formula can be expressed as follows:

$$\mathbb{E}(v_r) = \sum_{S_r} \frac{(M-1)!(T-M)!}{T!} \left(\frac{1}{M\hat{p}\left(v_t \mid S_r \cup v_r\right)} - \frac{1}{M\hat{p}\left(v_t \mid S_r\right)}\right) \qquad (6)$$

In the above formula, the factorial part is only related to the sampling number and the neighborhood node numbers, so $Q_T^M$ is used instead.

$$\mathbb{E}(v_r) = \sum_{S_r \subseteq N(v_r) \setminus v_t, |S_r| = M-1} Q_T^M \left(\frac{1}{\hat{p}\left(v_t \mid S_r \cup v_r\right)} - \frac{1}{\hat{p}\left(v_t \mid S_r\right)}\right) \qquad (7)$$

Besides, $Q_T^M$ is the same for all neighborhood nodes in $N(v_r)$ when calculating the respective cooperative causal weights under the condition that the sampling number is determined. Where $Q_T^M$ is denoted as follows:

$$Q_T^M = \frac{(M-1)!(T-M)!}{M \times T!} \qquad (8)$$

At this point, the cooperative causal weight taking the first-order neighborhood node as an example has been obtained, and the CoCa-sampling algorithm is used to guide the node sampling, the specific process is shown in algorithm 1.

---

**Algorithm 1** Cooperative Causal Sampling on First-order Neighborhood Nodes

---

**Input:** $G = (V, E)$, $N$, $v$, $n \in \{1, \ldots, N\}$, $M$

**Output:** Node set $C = \{v_1, v_2, \ldots, v_M\}$

1. **for** $n = 1, \ldots, N$ **do**
2.   find the neighborhood of $v_n$: $N(v_n)$
3.   **for** each $v_t$ in $N(v_n)$ **do**
4.     **if** $S_j$ in $N(v_n) \setminus v_t$ **then**
5.       calculate $\dfrac{1}{\hat{p}(v_t \mid S_r \cup v_r)} - \dfrac{1}{\hat{p}(v_t \mid S_r)}$
6.       calculate $\mathbb{E}(v_t) = \sum \left(\dfrac{1}{\hat{p}(v_t \mid S_r \cup v_r)} - \dfrac{1}{\hat{p}(v_t \mid S_r)}\right)$
7.     **end if**
8.   sampling nodes in $N(v_n)$ according to $E(v_t)$ : $C = \{v_1, v_2, \ldots, v_m\}$
9.   **end for**
10. **end for**

---

### D. Aggregation Functions and Parameters Learning

After conducting CoCa-sampling on neighboring nodes, a set $\mathcal{C}$ of sampling nodes is obtained. In order to aggregate the information of these nodes, this paper selected the mean aggregation function as the aggregation function of CoCa-GraphSAGE.

The mean aggregate function concatenates the $l-1$ layer representation vector of the target node with the feature of the nodes in its sampling set. Then the mean value is calculated for each dimension of the vector, and finally the $l$-order neighborhood representation of the node $v_r$ is obtained by nonlinear transformation. The mean aggregate function is expressed as:

$$\boldsymbol{h}_r^l \leftarrow \sigma(W^l \cdot \text{MEAN}(\{\boldsymbol{h}_{v_r}^{l-1}\} \bigcup \{\boldsymbol{h}_u^{l-1} : \forall u \in \mathcal{C}\})) \qquad (9)$$

Where $v_r$ is the target node, $l$ is the number of network layers, $l \in \{1, \cdots, L\}$ represents the $l$-order neighborhood, $\mathcal{C}$ is the set of neighborhood nodes obtained from CoCa-sampling, $u$ is the node in this set, $\text{MEAN}(\cdot)$ represents the mean operation function, $\sigma(\cdot)$ represents the nonlinear activation function.

In the parameter learning stage, CoCa-GraphSAGE is used for node classification tasks, so the cross entropy loss function is used to perform softmax normalization on the $l$-order node representation vector $\boldsymbol{h}_{v_r}^l$, and the normalized output vector is used as the vector obtained by forward propagation of the output, denoted as $z_r$.

$$z_r = \text{Softmax}(\boldsymbol{h}_{v_r}^L) = \frac{\exp(\boldsymbol{h}_{v_r}^L)}{\sum_{c=1}^{C} \exp(\boldsymbol{h}_{v_r,c}^L)} \qquad (10)$$

In the subsequent backpropagation stage, CoCa-GraphSAGE uses the Adam optimizer to optimize the parameters, for example, the parameters in the aggregation function. The classification loss function in the training stage is expressed as:

$$\mathcal{L} = -\sum_{r=1}^{N} \sum_{c=1}^{C} y_{rc} \log(z_{rc}) \qquad (11)$$

Where, $C$ represents the total number of node classes in the graph, $y_{rc}$ is the node label indicator variable. If $y_r = c$, then $y_{rc} = 1$, otherwise 0. $z_{rc}$ represents the $c$-th element in the node embedding representation vector $z_r$, representing the probability that node $v_r$ belongs to class $C$.

## IV. EXPERIMENTS

### A. Experiment Setup

In order to verify the performance of the proposed method on node classification tasks, five node classification task datasets are selected as the benchmark datasets in the experiment, which are as follows: There are three real citation network datasets (Cora, Pubmed, Citeseer), one co-author dataset (Coauthor-CS), and one large-scale graph reference dataset (ogbn-arxiv). The composition of each dataset is shown in Table I.

TABLE I
THE STATISTIC DETAILS OF DATASETS

| Datasets | Cora | Citeseer | Pubmed |
|---|---|---|---|
| Nodes | 2708 | 3327 | 19717 |
| Edges | 5429 | 4732 | 44338 |
| Classes | 7 | 6 | 3 |
| Features | 1433 | 3703 | 500 |
| Training nodes | 140 | 120 | 60 |
| Validation nodes | 500 | 500 | 500 |
| Test nodes | 1000 | 1000 | 1000 |

In this paper, GCN, GAT, GraphSAGE, and four improved models are selected as the comparison models of the experiment. The introduction and parameter Settings of the comparison models are as follows:

1) *GCN:* The model computes the embedding representation of nodes by learning the first-order nearest neighbors in spectral domain graph convolution. According to [28], the hyperparameters of GCN are set as follows: dropout rate=0.5, L2 regularization constant is $5\times10-4$, and the number of hidden units is 128.

2) *GAT:* The model obtains the embedded representation of nodes by introducing a multi-terminal attention mechanism in the first-order neighborhood. According to [7], the hyperparameters of GAT are set as follows: dropout rate=0.5, the number of network layers is 2, the first layer is set with 8 attention heads, the second layer is set with 3 attention heads, the number of hidden units is 128, and the $L2$ regularization constant is $5\times10^{-4}$.

3) *GraphSAGE:* The model learns the node embedding representation by uniformly sampling the neighborhood of each node and aggregating the information of its neighbors [5], the hyperparameters of GraphSAGE are set as follows: the number of neighborhood samples is $S1$=25, $S2$=10, the network layer is $K$=2, the number of hidden units is 128, and the aggregation function is chosen to use Mean.

4) *GraphSAGE with RL-based sampling (RL-GraphSAGE):* The model replaces random sampling in GraphSAGE with uniform sampling based on reinforcement learning. According to [8], the hyperparameters of RL-GraphSAGE are set as follows: discount rate=0.9, sampling number $S1$=25, $S2$=10, network layer $K$=2, and $L2$ regularization coefficient $5\times10^{-4}$.

5) *C-GraphSAGE:* The model uses causal sampling based on causal inference to aggregate information from neighborhood nodes. According to [13], its hyperparameters are set as follows: the learning rate is 0.01, the number of network layers $K$=2, the number of samples in the first layer is 25, the number of sampling nodes in the second layer is 10, the batch size is 50, the $L2$ regularization constant is $5\times10^{-4}$, and the aggregation function uses Mean.

6) *Graph Contrastive Learning with Negative Samples Selecting Strategy (GCNSS):* The negative sample sampling strategy is used to guide the selection of negative samples according to the learning results of some nodes [29]. The hyperparameters of GCNSS are set as follows: the learning rate is 0.01, the architecture is chosen to use GCN, and the negative sampling ratio is 0.6.

7) *Re-weight Nodes and Graph Learning Convolutional Network with Manifold Regularization (RN-GLVNMR):* The nodes are weighted adaptively according to the distance between the labeled node and the category margin [29]. Its hyperparameters are set as follows: the number of network layers $K$=2, the number of features of dimensionality reduction modules =70, and the number of features of hidden layer units =30.

8) *CoCa-GraphSAGE:* The hyperparameters are set according to the experimental experience: learning rate is 0.01, the number of network layers $K$=2, the number of sampling nodes in the first layer 10, the number of sampling nodes in the second layer 10, the batch size 50, and the $L2$ regularization constant $5\times10^{-4}$.

In order to verify the robustness of CoCa-GraphSAGE on the classification task, some experiments considering perturbations are set up. According to the existing studies, in the three datasets of Cora, Pubmed and Citeseer, Bernoulli matrix is used to add perturbation to the features of the citation network data set by means of the exclusive-OR operation, so as to evaluate the robustness of the model. At the same time, in order to further measure the influence of the strength of the perturbation, the perturbation ratio $\eta$ is defined:

$$\eta = \frac{1}{N \times D} \sum_{i=1}^{N} \sum_{j=1}^{D} \mathbf{X}_B(i,j) \qquad (12)$$

A higher perturbation ratio means a stronger perturbation is present in the data. In addition, for the Coauthor-CS and ogbn-arxiv data sets, probability-based additive Gaussian noise perturbation is introduced according to the above perturbation methods, and random Gaussian noise is first generated according to the variance of node features. Then, for each node, the probability of whether to add noise to it is selected by probability. For convenience of comparison, the probability is set to 0.1 to 0.5, which is defined as the perturbation ratio, indicating the proportion of disturbed nodes.

### B. Results and Analyses

#### 1) Classification Performance Experiments

In order to verify the performance of CoCa-GraphSAGE on node classification task, no perturbation experiments are carried out first. Table II shows the experimental results of each model. ▲ in the table represents the method proposed in this paper, and the performance indicators are displayed in the way of accuracy ± standard deviation %, and - indicates that the corresponding results are not provided in this methodology.

As can be seen from the data in Table II, compared with GraphSAGE and C-GraphSAGE, CoCa-GraphSAGE model shows slight performance improvement and better stability. This performance advantage stems from the use of CoCa-sampling in CoCa-GraphSAGE, which allows the model to more effectively identify and utilize neighborhood nodes that have a decisive impact on the predicted results. CoCa-GraphSAGE shows similar levels of performance compared to the GCN and GAT models. This means that CoCa-GraphSAGE adopts a CoCa-sampling method, which also realizes the effective use of neighborhood information. The slight outperformance of the GCNSS model over CoCa-GraphSAGE highlights the value of contrastive learning strategies. By directing the model to focus on differences between classification results, GCNSS facilitates more discriminative node representation learning and slightly improves classification performance. The RN-GLVNMR model performs slightly worse because its sampling strategy does not fully consider the complex dependencies between nodes. The above results show that CoCa-sampling strategy can effectively improve the accuracy of GraphSAGE in node classification tasks.

#### 2) Robust Performance Experiments

In order to evaluate the robustness of the model, perturbation experiments are conducted in this paper, and the perturbation experiments are divided into the following three parts:

1) In order to verify the overall robustness of each model, perturbations are added to both the training set and the test set;
2) In order to verify the robustness of each model in the training stage, perturbations are only added to the training set;
3) In order to verify the robustness of each model in the test stage, perturbations are only added to the test set.

Table III to Table VII summarize the classification performances of each model compared with other models when the perturbation ratio ranges from 0.1 to 0.5, and Fig. 5 to Fig. 9 show the corresponding trend graph of each dataset under the three perturbation addition modes. The red curve is the method proposed in this paper.

TABLE II
COMPARISON OF PERFORMANCE WITHOUT PERTURBATION

| Model | Cora | Pubmed | Citeseer | Coauthor-CS | ogbn-arxiv |
|---|---|---|---|---|---|
| GCN | 80.6±0.4 | 78.7±0.4 | 70.8±0.5 | 91.1±0.5 | 70.7±0.1 |
| GAT | 81.1±0.7 | 78.5±0.3 | 71.4±0.6 | 91.3±0.6 | 71.9±0.1 |
| GraphSAGE | 79.6±0.6 | 78.1±0.3 | 69.9±0.6 | 90.5±2.8 | 70.4±0.3 |
| RL-GraphSAGE | 80.9±0.3 | 79.4±0.2 | 71.2±0.4 | 90.2±0.8 | - |
| GCNSS | **81.4**±0.6 | **79.5**±0.4 | **71.7**±0.4 | 91.6±0.4 | - |
| RN-GLVNMR | 79.2±1.0 | 75.1±2.0 | 70.3±1.2 | 91.5±0.5 | - |
| C-GraphSAGE | 79.6±0.5 | 78.3±0.4 | 71.3±0.5 | 91.5±0.4 | 70.8±0.1 |
| CoCa-GraphSAGE▲ | 79.9±0.2 | 78.9±0.3 | 71.3±0.2 | **92.0**±0.3 | **71.8**±0.2 |

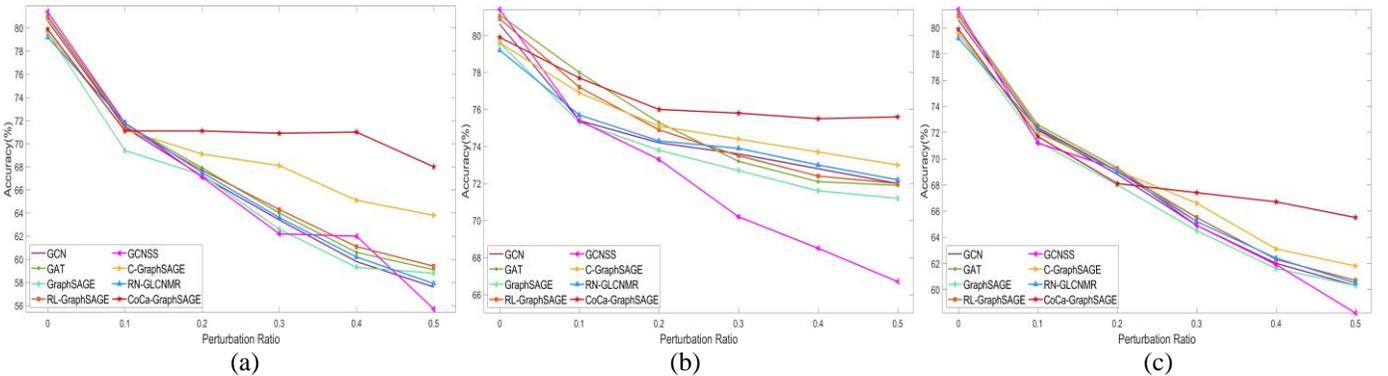

**Fig. 5.** Classification performance on Cora dataset with perturbation. (a) Test data perturbed, Train data perturbed, (b) Test data non perturbed, (c) Train data non perturbed.

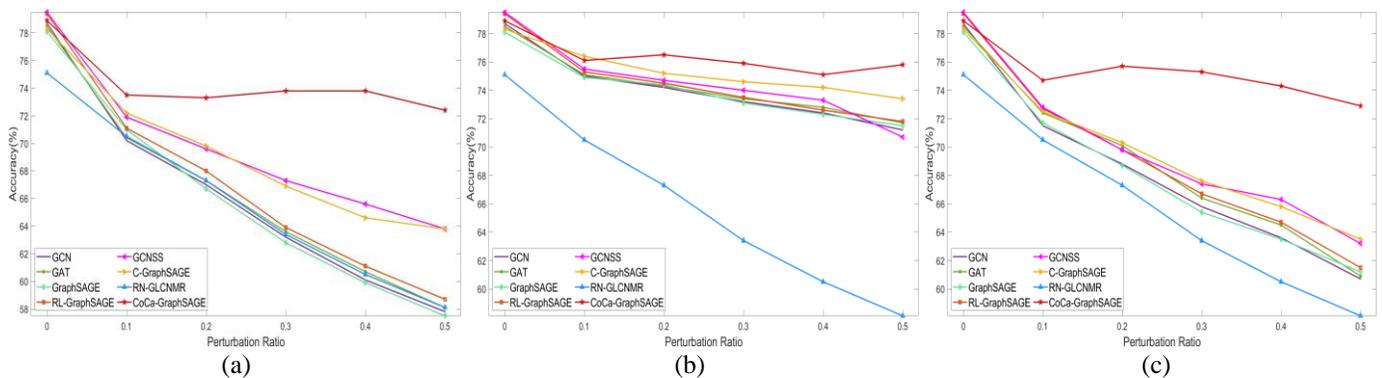

**Fig. 6.** Classification performance on Pubmed dataset with perturbation. (a) Test data perturbed, Train data perturbed, (b) Test data non perturbed, (c) Train data non perturbed.

TABLE III
CLASSIFICATION PERFORMANCE COMPARISON ON CORA WITH PERTURBATION

| Type of data | Model | Perturbation ratio | | | | |
|---|---|---|---|---|---|---|
| | | 0.1 | 0.2 | 0.3 | 0.4 | 0.5 |
| Test data perturbed | GCN | 71.4±0.3 | 67.2±0.5 | 63.4±1.8 | 59.8±2.2 | 57.6±1.8 |
| | GAT | 71.7±0.5 | 67.9±0.2 | 64.0±1.5 | 60.6±1.8 | 59.1±1.4 |
| | GraphSAGE | 69.4±0.6 | 67.3±0.3 | 62.6±0.7 | 59.3±1.1 | 58.8±1.7 |
| | RL-GraphSAGE | 71.5±0.4 | 67.7±0.4 | 64.3±0.9 | 61.1±1.3 | 59.4±1.1 |
| | GCNSS | **71.8**±0.5 | 67.1±1.1 | 62.2±0.6 | 62.0±1.1 | 55.7±3.5 |
| | C-GraphSAGE | 71.2±0.4 | 69.1±0.5 | 68.1±1.5 | 65.1±1.7 | 63.8±1.2 |
| | RN-GLCNMR | 71.8±1.1 | 67.5±1.3 | 63.6±1.9 | 60.2±1.8 | 57.9±1.9 |
| | CoCa-GraphSAGE ▲ | 71.1±0.2 | **71.1**±0.3 | **70.9**±1.1 | **71.0**±0.5 | **68.0**±1.6 |
| Test data non perturbed | GCN | 75.4±0.4 | 74.2±0.7 | 73.6±1.5 | 72.8±1.2 | 72.0±0.9 |
| | GAT | **78.0**±0.3 | 75.3±0.5 | 73.2±0.9 | 72.1±0.5 | 71.9±0.7 |
| | GraphSAGE | 75.3±0.5 | 73.8±0.4 | 72.7±0.7 | 71.6±1.3 | 71.2±0.9 |
| | RL-GraphSAGE | 77.2±0.6 | 74.9±0.9 | 73.5±1.1 | 72.4±0.7 | 72.0±1.2 |
| | GCNSS | 75.4±5.2 | 73.3±6.8 | 70.2±5.8 | 68.5±4.2 | 66.7±5.3 |
| | C-GraphSAGE | 76.9±0.3 | 75.1±0.3 | 74.4±0.5 | 73.7±0.7 | 73.0±1.0 |
| | RN-GLCNMR | 75.7±0.4 | 74.3±0.6 | 73.9±0.8 | 73.0±0.7 | 72.2±0.8 |
| | CoCa-GraphSAGE ▲ | 77.7±0.6 | **76.0**±0.2 | **75.8**±0.2 | **75.5**±0.7 | **75.6**±0.6 |
| Train data non perturbed | GCN | 72.2±0.7 | 68.8±0.7 | 64.9±0.7 | 62.0±1.1 | 60.3±0.5 |
| | GAT | **72.6**±0.3 | 69.3±0.9 | 65.2±1.0 | 62.4±0.8 | 60.5±0.4 |
| | GraphSAGE | 71.3±0.2 | 68.0±0.4 | 64.5±1.2 | 61.6±1.0 | 60.3±0.8 |
| | RL-GraphSAGE | 72.4±0.2 | 69.1±0.1 | 65.5±0.5 | 62.3±0.4 | 60.7±0.5 |
| | GCNSS | 71.2±0.2 | **69.2**±6.8 | 64.9±4.4 | 61.9±4.0 | 58.2±3.9 |
| | C-GraphSAGE | 72.0±0.4 | 69.1±0.5 | 66.6±0.8 | 63.1±0.4 | 61.8±0.3 |
| | RN-GLCNMR | 72.3±0.6 | 69.0±0.6 | 65.2±0.8 | 62.4±0.7 | 60.5±1.1 |
| | CoCa-GraphSAGE ▲ | 71.7±0.3 | 68.1±0.8 | **67.4**±0.5 | **66.7**±0.5 | **65.5**±0.6 |

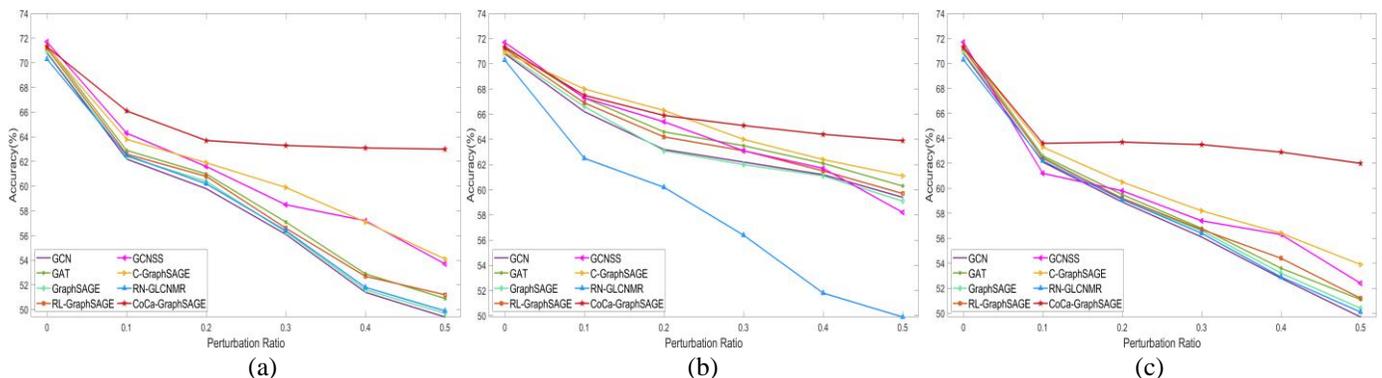

**Fig. 7.** Classification performance on Citeseer dataset with perturbation. (a) Test data perturbed, Train data perturbed, (b) Test data non perturbed, (c) Train data non perturbed.

TABLE IV
CLASSIFICATION PERFORMANCE COMPARISON ON PUBMED WITH PERTURBATION

| Type of data | Model | Perturbation ratio | | | | |
|---|---|---|---|---|---|---|
| | | 0.1 | 0.2 | 0.3 | 0.4 | 0.5 |
| Test data perturbed | GCN | 70.2±1.3 | 67.0±1.2 | 63.2±0.6 | 60.1±1.6 | 57.8±3.0 |
| | GAT | 70.4±1.1 | 67.3±0.5 | 63.6±1.6 | 60.7±0.9 | 58.1±2.4 |
| | GraphSAGE | 71.0±0.7 | 66.7±1.1 | 62.8±1.4 | 59.9±1.7 | 57.5±2.2 |
| | RL-GraphSAGE | 71.1±0.5 | 68.0±0.8 | 63.9±1.1 | 61.1±1.5 | 58.7±1.7 |
| | GCNSS | 71.9±0.5 | 69.6±0.8 | 67.3±1.2 | 65.6±0.9 | 63.8±0.9 |
| | C-GraphSAGE | 72.2±0.5 | 69.8±0.8 | 66.9±0.5 | 64.6±1.5 | 63.8±1.9 |
| | RN-GLCNMR | 70.5±0.8 | 67.3±0.7 | 63.4±0.9 | 60.5±0.9 | 58.1±1.1 |
| | CoCa-GraphSAGE▲ | **73.5**±0.5 | **73.3**±0.3 | **73.8**±1.3 | **74.8**±0.6 | **72.4**±0.6 |
| Test data non perturbed | GCN | 75.0±1.0 | 74.2±0.4 | 73.2±0.4 | 72.4±0.8 | 71.2±0.6 |
| | GAT | 75.1±0.7 | 74.3±0.5 | 73.4±0.3 | 72.8±0.3 | 71.7±0.8 |
| | GraphSAGE | 74.9±0.5 | 74.4±0.3 | 73.1±0.4 | 72.3±0.6 | 71.5±1.1 |
| | RL-GraphSAGE | 75.3±0.5 | 74.5±0.2 | 73.5±0.5 | 72.6±0.5 | 71.8±0.3 |
| | GCNSS | 75.5±5.3 | 74.7±4.0 | 74.0±3.6 | 73.3±1.0 | 70.7±1.4 |
| | C-GraphSAGE | **76.4**±0.4 | 75.2±0.4 | 74.6±0.3 | 74.2±0.7 | 73.4±0.2 |
| | RN-GLCNMR | 75.4±0.8 | 74.6±0.5 | 73.6±0.3 | 72.5±0.6 | 71.4±0.5 |
| | CoCa-GraphSAGE▲ | 76.1±0.5 | **76.5**±0.4 | **75.9**±0.5 | **75.1**±0.6 | **75.8**±0.3 |
| Train data non perturbed | GCN | 71.5±0.4 | 68.8±0.6 | 65.8±1.0 | 63.6±0.7 | 60.7±1.2 |
| | GAT | 72.4±0.8 | 70.1±0.5 | 66.4±0.4 | 64.5±0.4 | 60.9±1.6 |
| | GraphSAGE | 71.7±0.4 | 68.7±0.7 | 65.4±0.6 | 63.5±0.6 | 61.2±0.7 |
| | RL-GraphSAGE | 72.8±2.2 | 69.8±4.0 | 67.4±4.8 | 66.3±2.2 | 63.2±4.5 |
| | GCNSS | 72.8±1.0 | 71.9±1.4 | 69.5±3.3 | 65.3±3.6 | 62.5±1.9 |
| | C-GraphSAGE | 72.5±0.4 | 70.3±0.3 | 67.6±0.7 | 65.8±0.2 | 63.5±0.9 |
| | RN-GLCNMR | 71.7±0.5 | 69.2±0.5 | 66.0±0.8 | 63.9±0.4 | 60.9±0.6 |
| | CoCa-GraphSAGE▲ | **74.7**±0.6 | **75.7**±0.8 | **75.3**±0.5 | **74.3**±0.6 | **72.9**±0.6 |

TABLE V
CLASSIFICATION PERFORMANCE COMPARISON ON CITESEER WITH PERTURBATION

| Type of data | Model | Perturbation ratio | | | | |
|---|---|---|---|---|---|---|
| | | 0.1 | 0.2 | 0.3 | 0.4 | 0.5 |
| Test data perturbed | GCN | 62.2±0.4 | 59.8±0.7 | 56.1±0.4 | 51.4±0.9 | 49.4±1.3 |
| | GAT | 62.9±0.5 | 61.0±0.6 | 57.1±0.3 | 52.9±1.2 | 50.9±1.0 |
| | GraphSAGE | 62.4±0.4 | 60.4±0.8 | 56.3±1.2 | 51.6±1.0 | 49.7±1.5 |
| | RL-GraphSAGE | 62.6±0.7 | 60.8±0.3 | 56.6±0.9 | 52.7±1.4 | 51.2±1.1 |
| | GCNSS | 64.3±1.8 | 61.6±2.3 | 58.5±2.2 | 57.2±4.6 | 53.7±6.6 |
| | C-GraphSAGE | 64.1±0.2 | 61.9±0.4 | 59.9±0.7 | 57.1±0.5 | 54.1±1.2 |
| | RN-GLCNMR | 62.5±0.3 | 60.2±0.6 | 56.4±0.8 | 51.8±0.6 | 49.9±1.0 |
| | CoCa-GraphSAGE▲ | **66.1**±2.0 | **63.7**±2.0 | **63.3**±1.7 | **63.1**±1.5 | **63.0**±1.6 |
| Test data non perturbed | GCN | 66.2±0.6 | 63.2±0.3 | 62.2±1.0 | 61.2±0.6 | 59.4±0.3 |
| | GAT | 67.3±0.4 | 64.6±0.7 | 63.5±0.2 | 62.1±0.7 | 60.3±0.5 |
| | GraphSAGE | 66.6±0.3 | 63.1±0.7 | 62.0±1.0 | 61.1±0.5 | 59.1±1.1 |
| | RL-GraphSAGE | 66.9±0.2 | 64.2±0.8 | 63.1±0.4 | 61.5±1.1 | 59.7±0.5 |
| | GCNSS | 67.3±4.1 | 65.4±5.0 | 63.1±4.2 | 61.7±2.7 | 58.2±2.6 |
| | C-GraphSAGE | **68.0**±0.4 | **66.3**±0.6 | 64.0±0.4 | 62.4±0.2 | 61.1±0.5 |
| | RN-GLCNMR | 66.3±0.5 | 63.4±0.4 | 62.6±0.5 | 61.7±0.6 | 59.8±0.5 |
| | CoCa-GraphSAGE▲ | 67.5±0.3 | 65.9±0.6 | **65.1**±0.6 | **64.4**±0.6 | **63.9**±0.7 |
| Train data non perturbed | GCN | 62.1±0.5 | 58.9±1.0 | 56.1±0.7 | 52.8±0.5 | 49.7±1.4 |
| | GAT | 62.6±0.7 | 59.5±0.4 | 56.8±0.6 | 53.6±0.5 | 51.1±1.0 |
| | GraphSAGE | 62.5±0.4 | 59.1±0.3 | 56.6±1.1 | 53.2±0.6 | 50.4±1.1 |
| | RL-GraphSAGE | 62.4±0.2 | 59.2±0.5 | 56.7±0.8 | 54.4±1.0 | 51.2±0.6 |
| | GCNSS | 61.2±3.3 | 59.8±3.9 | 57.4±4.7 | 56.3±2.2 | 52.4±4.5 |
| | C-GraphSAGE | 63.3±0.2 | 60.5±0.6 | 58.2±0.5 | 56.4±1.2 | 53.9±0.8 |
| | RN-GLCNMR | 62.2±0.5 | 59.1±0.7 | 56.4±0.4 | 52.9±1.1 | 50.1±1.6 |
| | CoCa-GraphSAGE▲ | **63.6**±0.8 | **63.7**±0.3 | **63.5**±0.8 | **62.9**±0.7 | **62.0**±0.4 |

TABLE VI
CLASSIFICATION PERFORMANCE COMPARISON ON COAUTHOR-CS WITH PERTURBATION

| Type of data | Model | Perturbation ratio | | | | |
|---|---|---|---|---|---|---|
| | | 0.1 | 0.2 | 0.3 | 0.4 | 0.5 |
| Test data perturbed | GCN | 87.9±4.7 | 82.3±4.8 | 78.7±4.3 | 74.1±4.1 | 69.5±3.8 |
| | GAT | 86.1±1.7 | 83.1±1.1 | 79.8±2.1 | 76.4±3.4 | 74.0±2.1 |
| | GraphSAGE | 85.7±1.8 | 77.5±2.9 | 74.8±2.8 | 68.7±3.6 | 63.8±3.8 |
| | C-GraphSAGE | 87.8±1.7 | 84.7±1.1 | 81.4±2.1 | 77.9±3.4 | 75.4±2.1 |
| | CoCa-GraphSAGE ▲ | **90.1**±0.8 | **85.5**±1.5 | **81.9**±2.0 | **78.8**±1.5 | **75.7**±2.4 |
| Test data non perturbed | GCN | 87.9±2.8 | 85.5±3.1 | 82.6±3.3 | 80.1±4.1 | 78.8±3.6 |
| | GAT | 88.1±1.0 | 84.9±2.1 | 81.8±0.9 | 79.6±4.0 | 76.8±3.9 |
| | GraphSAGE | 90.4±0.5 | 86.8±0.6 | 83.5±2.5 | 80.7±2.4 | 80.1±2.0 |
| | C-GraphSAGE | 89.8±1.0 | 87.3±2.1 | 84.2±0.9 | 82.1±4.0 | 81.2±3.9 |
| | CoCa-GraphSAGE ▲ | **91.7**±0.3 | **90.1**±1.3 | **88.8**±1.0 | **87.6**±1.1 | **85.2**±1.4 |
| Train data non perturbed | GCN | 81.6±0.4 | 80.1±1.0 | 79.3±0.5 | 77.6±0.6 | 75.9±1.2 |
| | GAT | 89.7±1.3 | 84.6±1.3 | 80.4±1.4 | 76.5±2.1 | 74.3±2.8 |
| | GraphSAGE | 82.8±1.6 | 77.7±2.6 | 76.5±1.8 | 75.8±2.5 | 68.4±3.4 |
| | C-GraphSAGE | 89.6±1.3 | 82.7±1.3 | 79.7±1.4 | 77.4±2.1 | 75.2±2.8 |
| | CoCa-GraphSAGE ▲ | **90.0**±0.7 | **85.3**±2.0 | **83.0**±1.6 | **81.7**±1.8 | **77.5**±2.9 |

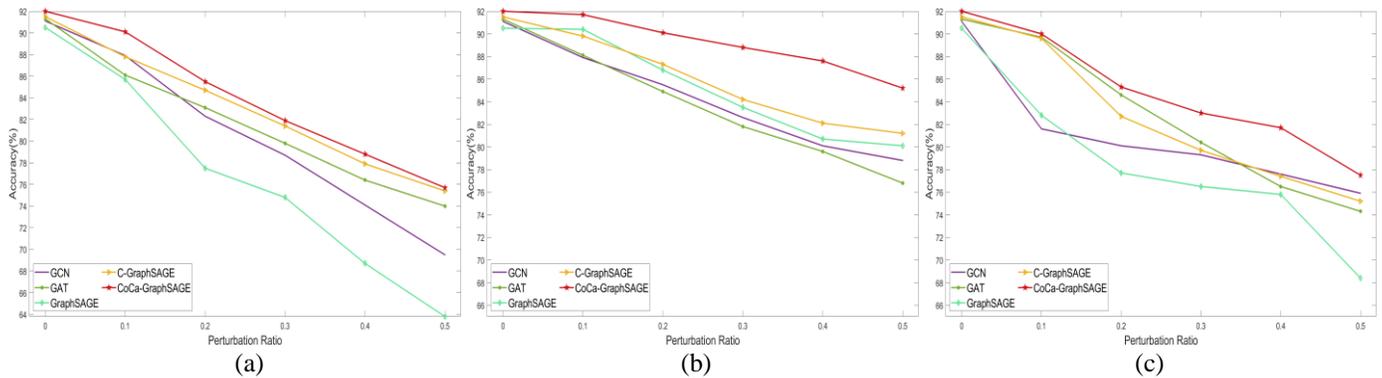

**Fig. 8.** Classification performance on Coauthor-CS dataset with perturbation. (a) Test data perturbed, Train data perturbed, (b) Test data non perturbed, (c) Train data non perturbed.

TABLE VII
CLASSIFICATION PERFORMANCE COMPARISON ON OGBN-ARXIV WITH PERTURBATION

| Type of data | Model | Perturbation ratio | | | | |
|---|---|---|---|---|---|---|
| | | 0.1 | 0.2 | 0.3 | 0.4 | 0.5 |
| Test data perturbed | GCN | 51.9±0.2 | 51.2±0.4 | 50.1±0.3 | 48.7±0.5 | 47.4±0.3 |
| | GAT | 52.4±0.4 | 50.6±0.3 | 49.9±0.2 | 48.2±0.3 | 47.8±0.2 |
| | GraphSAGE | 52.0±0.3 | 50.8±0.4 | 50.0±0.3 | 46.8±0.4 | 45.2±0.3 |
| | C-GraphSAGE | 54.9±0.5 | 52.4±0.5 | 51.5±0.6 | 50.4±0.7 | 50.9±0.6 |
| | CoCa-GraphSAGE ▲ | **57.0**±0.4 | **56.3**±0.5 | **55.4**±0.3 | **54.3**±0.4 | **53.7**±0.4 |
| Test data non perturbed | GCN | 54.4±0.3 | 53.7±0.4 | 51.9±0.5 | 51.1±0.6 | 50.2±0.4 |
| | GAT | 55.4±0.4 | 54.2±0.3 | 53.8±0.4 | 51.6±0.2 | 51.4±0.2 |
| | GraphSAGE | 52.5±0.4 | 52.2±0.5 | 52.1±0.6 | 51.4±0.3 | 50.7±0.5 |
| | C-GraphSAGE | 56.4±0.2 | 55.9±0.6 | 55.7±0.3 | 54.8±0.2 | 54.2±0.3 |
| | CoCa-GraphSAGE ▲ | **57.1**±0.3 | **56.9**±0.4 | **56.9**±0.5 | **56.5**±0.4 | **56.4**±0.5 |
| Train data non perturbed | GCN | 50.6±0.3 | 49.4±0.3 | 48.2±0.4 | 46.1±0.3 | 44.7±0.5 |
| | GAT | 50.9±0.4 | 48.6±0.4 | 47.7±0.3 | 46.3±0.4 | 45.6±0.4 |
| | GraphSAGE | 51.7±0.4 | 49.9±0.3 | 48.5±0.4 | 44.5±0.4 | 43.2±0.3 |
| | C-GraphSAGE | 53.4±0.4 | 52.8±0.2 | 50.8±0.3 | 48.6±0.4 | 46.9±0.5 |
| | CoCa-GraphSAGE ▲ | **57.3**±0.5 | **56.0**±0.4 | **55.2**±0.4 | **54.1**±0.7 | **52.8**±0.5 |

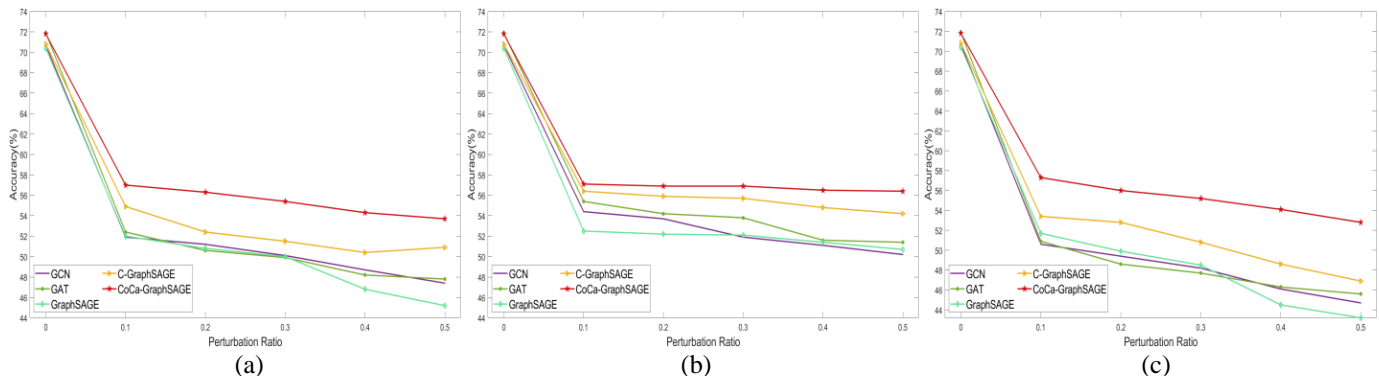

**Fig. 9.** Classification performance on ogbn-arxiv dataset with perturbation. (a) Test data perturbed, Train data perturbed, (b) Test data non perturbed, (c) Train data non perturbed.

From the data in Fig. 5 to 9, it can be seen that CoCa-GraphSAGE outperforms the comparison model when there are perturbations. Compared with GraphSAGE, CoCa-GraphSAGE shows the greatest improvement in test accuracy of 11.7%, 14.9%, 13.3%, 11.9% and 8.5% on the five data sets at the same perturbation ratio, respectively. Compared with C-GraphSAGE, CoCa-GraphSAGE shows the greatest improvement in test accuracy of 5.9%, 10.2%, 8.9%, 5.5% and 3.9% on the five datasets with the same perturbation ratio, respectively. These significant performance improvements demonstrate the effectiveness of CoCa-sampling in identifying robust node combinations.

When the perturbation ratio is lower than 0.2, the accuracy of the models is similar. This indicates that although the performance of the models is degraded by noise, the models have similar capabilities in dealing with minor perturbations due to the low perturbation ratio and fewer affected nodes. It is shown that the main competitiveness between models in a mildly disturbed environment comes from their ability to process the original feature information, rather than their robustness to the perturbation.

With the increase of the perturbation ratio, the classification accuracy of GCN, GAT and GraphSAGE decreases significantly, and the unknown perturbations in the data affect the feature attributes and importance evaluation of nodes, while the attention mechanism and random sampling mechanism lack robustness to noise. In the three datasets of Cora, Pubmed and Citeseer, although RL-GLCNMR has a high classification accuracy when there is no perturbation, and still maintains the advantage of stable standard deviation compared with GraphSAGE when considering perturbations, its classification accuracy also shows a significant decrease. There is no obvious difference with GraphSAGE, and the data-driven sampling strategy it uses has weak ability to distinguish noisy nodes. The performance of RN-GLCNMR is only slightly better than that of GCN and other classical methods, and it also fails to demonstrate excellent robustness, indicating that its strategy of adjusting the influence of nodes according to the distance from nodes to edges is sensitive to the perturbation of nodes themselves. In contrast, GCNSS, C-GraphSAGE and CoCa-GraphSAGE have relatively high accuracy when considering perturbations. The classification prediction guide node sampling strategy used by GCNSS can judge the validity of nodes in the process of message passing, so it has a certain robustness to perturbations. However, the fact that the standard deviation of GCNSS in all cases is much higher than the other limits indicates that the performance improvement brought by the scheme it uses is not stable. From this point of view, the increased robustness brought about by causation has a clear advantage, especially CoCa-GraphSAGE, which has consistently maintained excellent performance in all three datasets. The reason is that the CoCa-sampling used by CoCa-GraphSAGE is to analyze the causal effect in the case of node combination, which can effectively find the robust node combination. Although C-GraphSAGE also has remarkable robustness, it is limited by the weight bias of causal sampling, and its accuracy is lower than CoCa-GraphSAGE when the perturbation ratio is high. This result confirms the previous analysis of the causal sampling weight bias, and only single node causal analysis cannot fully learn the causal information on the graph. This means that the process of node information aggregation is not only dependent on some nodes with strong causal effect, but also closely related to the cooperation between neighboring nodes, which also echoes the theme of GraphSAGE to complete the goal of node information aggregation. The performance advantage of CoCa-GraphSAGE on the Coauthor-CS and ogbn-arxiv datasets also fully illustrates this principle, and it can be seen that CoCa-sampling can obtain more robust classification results than causal sampling.

### 3) Sample Node and Time Consumption Experiments

In addition, in CoCa-sampling, the number of sampling points, as a hyperparameter, also plays an important role in the performance of the model, which limits the selection of the feature information of the neighborhood nodes. In order to verify the influence of different sampling points on improving the robustness of the model, different sampling points are selected for experiments when perturbations exist in both training data and test data and the perturbation ratio is 0.5. Cora is selected as the experimental dataset. Fig. 10 shows the performance of CoCa-GraphSAGE and the running time of each epoch of the model.

Through comparative analysis of the performance of the model under different number of sampling nodes, it can be found that the accuracy of the model under the same perturbation ratio is gradually improved with the increase of the number of sampling points, and the accuracy is increased from 60% to 72%. When the number of sampling nodes is small, each additional bit of data will improve the learning ability and robustness of the model.

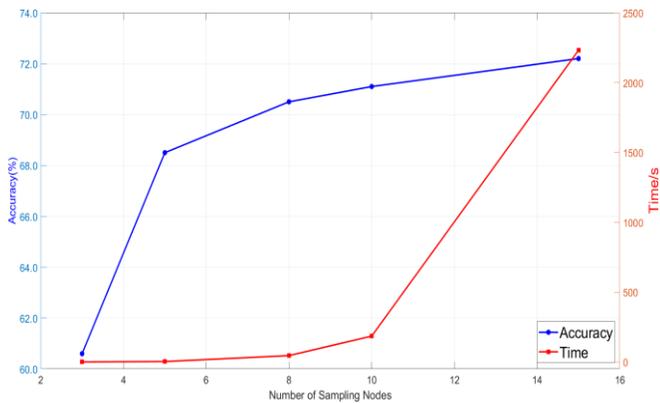

**Fig. 10.** The accuracy and running time of a single epoch under different number of sampling nodes in Cora dataset.

However, as the amount of data increases, the gradient of the curve representing the accuracy in the graph gradually decreases especially after the number of sampling nodes reaches 8, implying that the performance improvement of the model does not always continue to increase significantly as the dataset improves. It shows that although the number of sampling points can continue to increase, the model learns less and less useful information from each novel data point, and the model has captured most of the information. It also shows that CoCa-sampling can prioritize nodes with strong causal attributes in the neighborhood nodes, and these nodes bring high performance payoffs.

In addition, CoCa-GraphSAGE also has a learning capacity limit, that is, the limit of the complexity of the data it can learn and represent, and once this limit is reached, adding more data does not significantly improve the performance of the model, and may even lead to overfitting. The time consumption curve on the right reflects the negative effect brought by the increase in the number of sampling nodes: the running time of the model also starts to increase sharply after reaching 8, which is similar to the experimental result of the number of sampling nodes in GraphSAGE. Therefore, in practical applications, it is a key to find the optimal number of sampling points to balance robustness and time consumption.

## V. CONCLUSION

This paper proposes a novel model called CoCa-GraphSAGE, which effectively addresses the issue of node weight bias in random sampling and causal sampling by integrating two methods, cooperative game theory and Causal GraphSAGE. CoCa-GraphSAGE takes into consideration the cooperative relationships between neighboring nodes, which has been overlooked in previous researches. The model applies the cooperative causal sampling algorithm to partition the cooperative alliances within the neighboring nodes, constructing a cooperative causal model to analyze causal relationships across different cooperative alliances. It further calculates the cooperative causal sampling weights of candidate neighbors. Experimental evaluations conducted on five publicly available datasets demonstrate that the proposed CoCa-GraphSAGE performs comparably to the baseline model without perturbation and outperforms the baseline model in the presence of perturbations, indicating its outstanding robustness.

However, there are still some limitations in CoCa-GraphSAGE. It exhibits limited performance improvements without the introduction of perturbations. Additionally, when dealing with large-scale graphs, the cost of time may increase, and excessively restricting the sampling nodes could potentially result in information loss. In the future work, utilizing cooperative game theory to further improve the performance of sampling may be a valuable research direction.